# Wholesale Electricity Price Forecasting using Integrated Long-term Recurrent Convolutional Network Model

Vasudharini Sridharan, *Student Member*, Mingjian Tuo, *Student Member, IEEE* and Xingpeng Li, *Member, IEEE*

*Abstract*— Electricity price is a key factor affecting the decision-making for all market participants. Accurate forecasting of electricity prices is very important and is also very challenging since electricity price is highly volatile due to various factors. This paper proposes an integrated long-term recurrent convolutional network (ILRCN) model to predict electricity prices considering the majority contributing attributes to the market price as input. The proposed ILRCN model combines the functionalities of convolutional neural network and long short-term memory (LSTM) algorithm along with the proposed novel conditional error correction term. The combined ILRCN model can identify the linear and non-linear behavior within the input data. We have used ERCOT wholesale market price data along with load profile, temperature, and other factors for the Houston region to illustrate the proposed model. The performance of the proposed ILRCN electricity price forecasting model is verified using performance/evaluation metrics like mean absolute error and accuracy. Case studies reveal that the proposed ILRCN model is accurate and efficient in electricity price forecasting as compared to the support vector machine (SVM) model, fully-connected neural network model, LSTM model and the LRCN model without the conditional error correction stage.

*Index Terms*— Convolutional neural network, Deep learning, Energy price forecasting, Locational marginal price, Long short-term memory, Long-term recurrent convolutional network, Real time market price, Wholesale power energy market.

## I. INTRODUCTION

Wholesale energy market price, also known as settlement point prices or dynamic tariff, was first introduced in the year 1980's [1]. When electricity market is compared with other commodities, power trade exhibits multiple attributes: constant balance between production and consumption [2]; dependence of the consumption/load with respect to time, e.g. hour of the day, day of the week, and time of the year or whether it's a weekend or weekday or public holiday; load and generation that are influenced by external weather conditions [3], participating distributed energy resources [4], demand response program; and influence of neighboring markets [5] and other contributing factors like fuel price.

The dynamics of power prices have become very complicated, i.e. continuously varying price with sudden unexpected peaks as shown in [3]. In a deregulated power market such as locational marginal pricing (LMP) based market, the power prices are significantly influenced by the above-mentioned attributes.

Generally, wholesale electricity prices are likely to be high during peak demand periods and be low during off-peak demand periods [6]-[7]. The dynamic tariff or prices is an inherent load management method for properly allocating resources, thus ensuring the overall economic reliability [8]. Additionally, electric tariff is widely utilized as a fundamental control signal to support the demand response management for improving energy efficiency [9] and relieving the load burden on the power grid. Deep learning (DL) algorithms can capture nonlinear variations and the spatio-temporal patterns in energy market prices and demand.

Accurate forecasting of electricity prices is very important for each market participant. Electricity price has direct influence on the behavior of individual customers, distributed energy resource aggregators and local microgrid operators that aim for maximum profit or minimum cost. Electricity price affects their energy consumption profile and the amount of power sell to or purchase from the power grid. Accurate price predictions are essential for them to make informed decisions like demand scheduling and power dispatching, which may also enhance the reliability of bulk power systems. However, it is very challenging to predict electricity prices since they are highly volatile because of unexpected peaks and troughs, continuously altering supply and demand fluctuations, various system constraints, and other factors over the course of the day.

To address this issue, this paper is focused on wholesale market electricity price forecasting with the proposed integrated long-term recurrent convolutional network (integrated LRCN, or ILRCN) model. The proposed ILRCN model consists of a hybrid neural network architecture, i.e. LRCN, with an additional stage of conditional error correction [10]. In this paper, the Electric Reliability Council of Texas (ERCOT) wholesale market price and its contributing factors such as load and weather conditions for the Houston region of Texas are considered to demonstrate the proposed ILRCN model. In addition, this paper also verifies the practicality and feasibility of the proposed electricity price forecasting algorithm compared to existing algorithms.

The rest of the paper is organized as follows. Background analysis, challenges encountered in price forecasting and current industrial trends and models are covered in section II. Section III introduces the fundamentals of wholesale power energy market price and various neural networks, as well as a naive time-delayed price prediction method. Section IV describes the proposed ILRCN model and the overall methodology. Section V provides the simulation results and analyzes the performance of the proposed method. Section VI concludes the paper.

Vasudharini Sridharan, Mingjian Tuo and Xingpeng Li are with the Department of Electrical and Computer Engineering, University of Houston, Houston, TX, 77204, USA (e-mail: vsridharan2@uh.edu; mtuo@uh.edu; Xingpeng.Li@asu.edu). Sridharan Vasudharini Sridharan and Mingjian Tuo are co-first authors; the corresponding author is: Xingpeng Li.

## II. Literature Review

Accurate models for predicting wholesale electricity prices are necessitated due to the fact that electricity price is fundamental input to energy companies' decision-making mechanisms at the corporate level. A power utility company or large industrial consumer who is able to forecast the volatile wholesale prices with a reasonable level of accuracy can adjust its bidding strategy and its own production or consumption schedule in order to reduce the risk or maximize the profits in power trading. The market price depicts the substantial value of electric power.

The fluctuations in the supply and demand determine the rate of change of electricity price. Reference [11] shows how the power price varies in accordance with these fluctuations can reflect the actual value of electricity in the transaction process. Bidding procedures and settlement point price determine the profit level for participating power companies. Higher accuracy of power price prediction could enable participants to make better decisions that lead to higher profits or lower risks. There are various factors contributing to electricity price prediction: (a) quantifiable factors such as historical and recent electricity price and loads and (b) non-quantifiable factors such as market design and network topology. All these factors greatly increase the difficulty of electricity price prediction [12].

Presently, there are various studies related to electricity price forecasting [13]. Electricity price is utilized as a basic control signal to support the demand response management which is a feasible solution for improving energy efficiency. Price forecasting benefits power grid as it offers specific price instructions for participants to manage their flexible power usage at different time, which alleviates the load burden of power grid especially in peak demand time. Additionally, price forecasting also encourages consumption by end users during periods of valley demand along with reduced price and also allows customers to have multiple choices to determine the period of peak consumption. Amjady [14] explained the need for short-term electricity price forecasting and proposed models for such predictions.

A drawback of various statistical models proposed previously in the literature like auto-regressive, dynamic regression and transfer function is the fact that they use linear forecasters; hence as such, they do not perform well where the periodicity of input data is high, e.g. hourly data with rapid variations, i.e. the nonlinear behavior of hourly prices might become too complicated to predict [15]. But these models perform adequately if the data frequency is low, e.g. weekly/monthly patterns with small variations. To address the issues in statistical models and predict the nonlinear characteristic of electricity prices, different machine learning methods have been proposed. Among them, multilayer perceptron's (MLPs), support vector regressors and radial basis function networks are the most widely used.

Over the last decade, the field of neural networks has undergone several shifts that have led to what is known as deep learning today. The major issue of neural networks had always been the heavy computational complexity and computational cost of training large models. However, this issue can be addressed; for instance, in [16], a deep neural network is trained efficiently using an algorithm called greedy layer-wise pre-training or batch training process. As related developments followed, researchers were able to efficiently train complex neural networks whose depth was not just limited to a single hidden layer (as in the traditional multilayer perceptron).

In [17], a hybrid model with the combined kernel function is used for price forecasting of Australian electricity market. [18] uses generalized mutual information for normalization of input data and uses least square support vector machine for day-ahead price forecasting. Reference [19] also employs a hybrid algorithm for load and price forecasting along with simultaneous prediction of peak hour load and day-ahead price, applied on different datasets of NYISO and PJM accordingly. This was further being used to optimize demand side management in [20]. Deep neural network [21], gated recurrent unit [22], shallow neural network and long short-term memory (LSTM) [23] models are efficiently forecasting methods which aid in increasing their prediction accuracy compared to traditional statistical models. However, the benefits of these models are not exhibited completely due to continuous fluctuation in the market price.

Other energy related applications and use-cases show excellent results obtained in time series prediction [24]-[25], where the electricity price prediction is possible by using DL architectures. In reviewing the above details, we focus our research on modelling the wholesale electricity price forecasting using the proposed ILRCN model which can identify the non-linear behavior of market price due to various uncertainty factors contributing to energy price.

## III. Preliminaries

In this section, theoretical concepts of deep learning and fundamentals of electricity price are introduced.

### A. Electricity Market Price

There are two main types of electricity market price: day-ahead market (DAM) price and real-time market (RTM) price [26]. In day-ahead market, bids are submitted for interval hours of the operating day one day in advance, while in real-time market bids are submitted only a couple of hours in advance. These bids are highly dependent on several factors including the demand for the particular interval, demand response operating for the area, weather conditions, bidding strategies for the participating players. These bids are defined per interval, i.e. every market player can submit bids or use default bids. Market operators use the collected bids to compute the market settlement point price for each interval, as well as the generation scheduling.

Locational marginal pricing is used to price energy on the market in response to changes in supply and demand and the system's physical constraints. LMP accounts for the cost to produce the energy, the cost to transmit this energy within regions of the market, and the cost due to power losses as the energy is transported across the system.

### B. Naive Method

The naive method implemented in this paper serves as a benchmark to gauge the proposed ILRCN model as well as other price prediction models. It is a time-delayed price determination method that (i) simply takes the electricity prices of day *d-1* as the forecasted electricity prices of day *d* when conducting real-time price prediction one day ahead, or (ii) simply takes the electricity prices of hour *t-1* as the forecasted electricity prices of hour *t* when conducting real-time price prediction one hour ahead. This naive method works best for short term price forecasting (predictions over short look-ahead period such as 15 minutes), where the electricity price has relatively less volatility and is unlikely to change substantially during consecutive short time intervals.

### C. Support Vector Machine

Support vector machine (SVM) is a supervised type pf machine learning algorithm which is based on Statistical Learning Theory. Due to its greater ability in generalizing the problem, SVM has been successfully applied in classification tasks and regression tasks, especially on time series prediction and financial related application.

When using SVM in regression tasks, the Support Vector Regressor (SVR) uses a cost function to measure the empirical risk [27]. For given training data $\{(x_1, y_1), ..., (x_n, y_n)\} \subset \chi \times \mathbb{R}$, where $\chi$ denotes the space of the input data. Basically, the goal of regression is to find the function $f(x)$ that best models the training data, and at the same time as flat as possible. The case of linear function $f$ can be described as,

$$f(x) = \langle w, x \rangle + b, \quad w \in \chi, b \in \mathbb{R} \quad (1)$$

where $\langle \cdot, \cdot \rangle$ denotes the dot production in $\chi$. To construct the optimal hyper plane which can be used for regression, the problem can be reformulated as a convex optimization problem:

$$\text{minimize} \quad \frac{1}{2}\|w^2\| \quad (2)$$

$$\text{subject to} \quad \begin{cases} y_i - \langle w, x_i \rangle - b \leq \varepsilon \\ \langle w, x_i \rangle + b - y_i \leq \varepsilon \end{cases} \quad (3)$$

Constraint (3) ensures the convex optimization problem is feasible. In this study, SVR model is built based on the historic data to subsequently predict the electricity market. The goal of SVR training is to minimize the sum of squares error on the training set.

### D. Deep Neural Network

Deep learning brings about an explosion of data in all forms across the globe. The multilayer perceptron is a fully connected neural network (FCNN) architecture with one input layer, one or multiple hidden layers, and one output layer. In FCNN, all the neurons of the previous layer are fully connected to the neurons of the next layer as shown in Fig. 1.

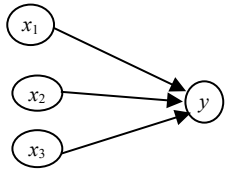

**Fig. 1.** Architecture of a sample two-layer deep learning model.

The basic DL model, an FCNN model, is the extension of the traditional MLP that uses multiple hidden layers. Each layer consists of a defined number of neurons. In Fig. 1, $x_1$, $x_2$, $x_3$ are the input neurons of the previous layer and $y$ is the neuron of the next layer which is the output in this scenario. The $y$ can be calculated by the following computation equation,

$$y = \sum_i \omega_i * x_i + b \quad (4)$$

where $\omega_i$ denotes the weights of the neurons and $b$ is the bias value; $x_i$ is the input neuron. A basic DL model would require activation function to be trained efficiently where each neuron $x_i$ is activated in each layer after the computation of $y$ in the preceding layer.

### E. Convolutional Neural Network

The convolutional neural network (CNN) [28]-[30] utilizes the concept of weight sharing which provides better accuracy in highly non-linear problems such as power price forecasting. It's an expansion from basic deep learning model. An example of a convolutional neural network is shown in Fig. 2 [31].

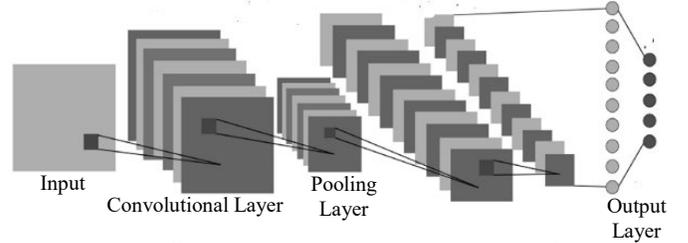

Fig. 2. Convolutional neural network architecture [31].

CNN consists of different types of hidden layers:
1) *Convolutional Layer*: The convolutional layer operates on two signals – (a) input and (b) filter (kernel) on the input. The underlying process here is the matrix multiplication of the input set and kernel to get the modified input, which extracts the information from the entire input using the kernel to obtain the essential data.
2) *Pooling Layer*: It is a sample based discretization process. It aims at reducing dimensionality of an input feature (e.g., input data, feature extraction) and extracting the information about the relation between input and output contained in the subset.
3) *Output Layer*: It is also known as fully-connected layer that connects every neuron in one layer to every neuron in subsequent layer. The output layer is also illustrated in Fig. 2.

### F. Long Short-Term Memory Neural Network

Recurrent neural network (RNN) is another framework of DL which uses the internal state to process a sequence of inputs [32]. Long short-term memory is an extended framework of RNN which can exhibit temporal behaviour of time-series input data. LSTM is capable of learning the long-term dependencies from the time sequential data such as electricity prices. The fundamental equations of an LSTM network can be represented as follows:

$$h_t = (1 - z_t) * h_{t-1} + z_t * h_t \quad (5)$$
$$z_t = \sigma(W_f[h_{t-1}, x_t] + b_f) \quad (6)$$

where, $x_t$ is the network input; $h_t$ is the output state of the neuron from LSTM network; $h_{t-1}$ is the previous state of the neuron; $z_t$ computes the necessary information and removes the irrelevant data; $\sigma$ is the sigmoid function; $W_f$ is the weight function and $b_f$ is the bias value.

### G. Nonlinearities in Electricity Prices

LMP is the basis of wholesale power market mechanisms for many nations including the United States. LMP is affected by a number of influencing factors including demand, availa-ble supply, network topology and losses, system constraints, and climatic conditions. The congestion component of LMP accounts for the network constraints that bind the optimal generating strategy. Overall, LMP could vary substantially for different locations and different time intervals and it is highly non-linear.

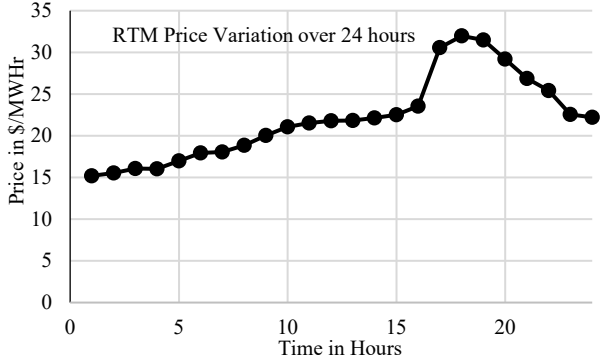

Fig. 3. Variation of settlement point price with respect to time in hours.

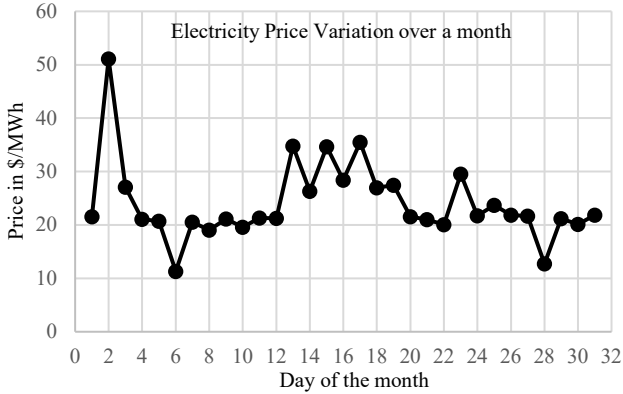

Fig. 4. Variation of settlement point price over a month.

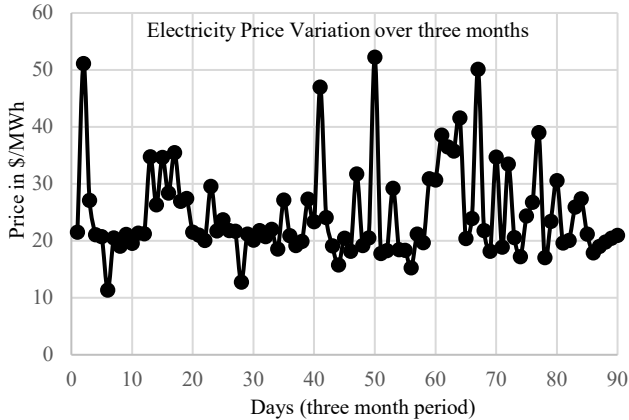

Fig. 5. Variation of settlement point price over three months.

Figures 3, 4 and 5 show the variation of ERCOT prices with respect to different time horizons. Fig. 3 displays the RTM settlement price variation over the course of the day showing the non-linear and continuous fluctuation of electricity price. According to Fig. 4 and Fig. 5, it is evident that the price dynamics have various seasonal patterns, corresponding to a daily and weekly periodicity, and are also influenced by calendar effect, i.e. weekdays *versus* weekends and holidays.

These properties also apply to the demand. The essential component for accurate price prediction is understanding the variation of price, i.e. identifying peaks and troughs and its relation to its governing factors contributing to change in price.

IV. SYSTEM MODEL

The price forecasting is classified into short term, medium term, and long term forecasting [33]. The short term ranges from one hour ahead to several hours ahead; a few hours to 1 week ahead forecasting is medium term forecasting; and beyond that it is the long term forecasting. We focus on hour-ahead and day-ahead forecasting in this work. The proposed forecasting model can be formulated as:

$$MP_t = F_t + E^*_{t-1} \qquad (7)$$

where, $t$ is the current time interval; $MP_t$ is the forecasted market price by the proposed ILRCN model; $F_t$ is the price forecasted by the hybrid neural network model, LRCN, that is explained in section IV.B; and $E^*_{t-1}$ is the electricity price forecasting error correction component from the previous time interval and is also referred to as the calibrated value that is defined in Section IV.C.

The $E^*_{t-1}$ aims to reduce the forecasting error that is partially due to insufficient details/features contributing to energy price in the dataset under consideration. It can consider the prediction errors of a few previous time intervals.

Fig. 6 illustrates the flowchart of short-term electricity price forecasting using the proposed ILRCN model based on several prior hours of input data. The historical data is the input forming the initial stage for the establishment of the proposed ILRCN model. The linear and non-linear behaviour and characteristics of the input data are analysed using the proposed ILRCN model. Input feature pre-processing is used to normalize, scale, and define the features from the input data to improve the accuracy of price forecasting. In order to tune parameters of the model, optimization and hyper-parameter tuning method with cross validation is utilized along with the LRCN model. Finally, the forecasted price coming from the proposed LRCN model is adjusted to incorporate the conditional error correction component; then, the desired output is obtained in this study.

The detailed explanation of the various components of the proposed ILRCN model is introduced in detail in the following subsections.

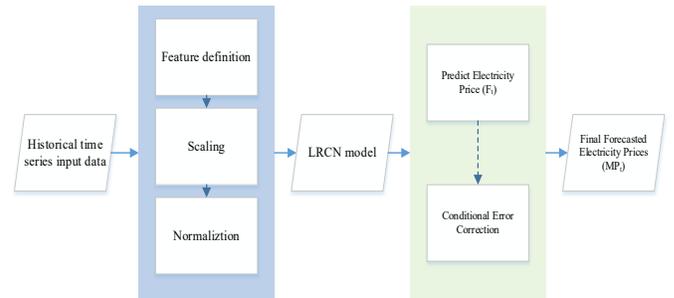

Fig. 6. Flowchart of short-term electricity price forecasting with ILRCN.

A. *Feature Definition and Preprocessing*

Electricity price is mainly affected by electrical demand which varies over time depending on season, weather and generation cost [34]. In order to forecast the market electricity price, date of delivery, temperature, and load data are selected as input features. The input data are normalized using various scaling functions to avoid excess stress on a specific dimension during the training process. For the date of delivery data which has no ordinal relationship, one-hot encoding is applied, and min-max scalar technique is used to bound the temperature values within [0, 1]. After the normalization process, all major features contributing evidently to the electricity price are selected and defined to avoid over-fitting during the training process. After feature selection process, the processed data are converted into a moving windowed dataset, where preceding interval data and features are the inputs to forecast the current output which is the electricity price of current interval.

### B. Long-term Recurrent Convolutional Network

The artificial neural network is a non-linear model, which is capable of making accurate prediction for forecasting problems mentioned in [3]. The main parameters of the neural network model are the number of input vectors, the number of layers and the number of neurons in each layer [16]-[17], [20]. However, the large and sudden spikes in the input data evident from Figures 3, 4 and 5 will lead to less accuracy in the output. To mitigate the impact of the price outliers, an error correction stage is considered in this study; the proposed conditional error correction term can partially account for unavailability of information in the input dataset under consideration contributing to sudden variation in price.

Based on (4) to (6), the neural network model is employed to capture the relationship and the linear and non-linear behaviour within the input data. After input processing stage, the processed data, and the best subset of parameters containing the information obtained from input features are used in the proposed ILRCN model to forecast electricity price.

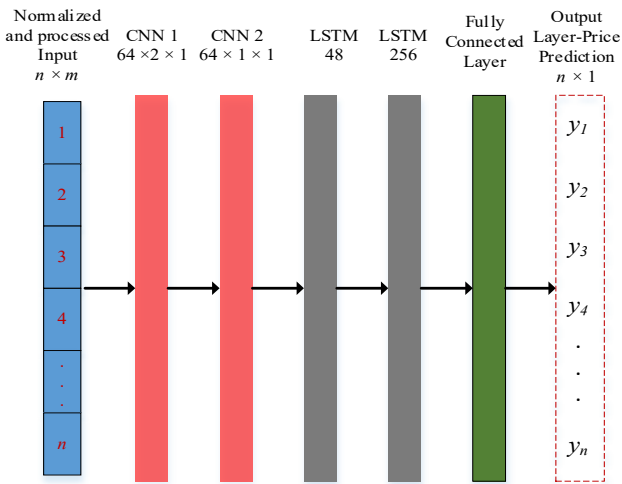

Fig. 7. The architecture of the proposed LRCN model.

Fig. 7 shows the structure of the proposed LRCN model, a hybrid neural network, which corresponds to the third block in Fig. 6. The proposed LRCN model includes two 1D convolution layers in the CNN segment, which aims to process spatial features and improve training efficiency. Two LSTM layers are added after the second convolution layer in the model to capture the dependencies in the sequential data. LSTM shows an improved performance for time-series data when compared to traditional models such as statistical forecasting model and regression model. In general, rectified linear unit (ReLU) is a widely used activation function as described in (8).

$$ReLU = \max(0, x) \quad (8)$$

where $x$ is the input features mapped as neuron.

This hybrid neural network can be used to build a function to express the variation among the historical data. Assuming the first input data $H$ of the total input data set consist of $n$ preceding hours' wholesale electricity price data, $H$ can then be formulated as:

$$H = \{D_1, D_2, \ldots, D_i, \ldots D_n\} \quad (9)$$

where $D_i$ represents the interval input data that is $i$-th interval prior to the current interval and $i$ ranges from 1 to 24.

Henceforth, $D$ can be expressed as:

$$D = \{y_1, y_2, \ldots y_j, \ldots y_m\} \quad (10)$$

where $y_j$ represents the features of the input set such as temperature, load, and date.

### C. Conditional Error Correction Term

It is unlikely that the LRCN model can predict the outlier prices with high accuracy. The forecasting error can be substantial when there are great differences among the input data sets due to non-linearities and spikes in the input data and when there are not sufficient information available for accurate prediction. To address this issue, we propose a novel conditional error correction term that is added to the price forecasted by the LRCN model to establish the price prediction.

The predicted value of the electricity price from the model is compared with the actual value of the settlement point price obtained from ERCOT. Then, the calibrated value $E_{t-1}^*$ from (7) accounting for insufficient input features contributing to electricity price is formulated as follows:

$$E_{t-1}^* = P_{t-1} - F_{t-1} \quad (11)$$

where $P_{t-1}$ denotes the actual price of time interval $t$-1; $F_{t-1}$ is the forecasted price of the same time interval $t$-1 by the proposed ILRCN model.

Since the LRCN model can well handle regular electricity prices, the proposed conditional error correction term will only be applied in the scenarios when the electricity price is very high and the error of price forecasting in the previous time period is beyond the pre-specified threshold.

### D. Metrics

The accuracy percentage of price forecasting models is defined in (12). It is very unlikely to predict the exact price; as a result, the number of correctly predicted price in (12) is counted with an error threshold that is the maximum acceptable deviation of predicted price from its actual value. The error threshold ranges from $1 to $3 in this paper.

$$Accuracy\ Percentage = \frac{No.\ of\ correctly\ predicted\ prices}{Total\ no.\ of\ test\ data\ points} \quad (12)$$

The mean absolute error (MAE) represents the average of the absolute difference between the actual and predicted values in the dataset [35]. The mean absolute error for the training process is calculated as follows:

$$MAE = \frac{1}{N}\sum_{i=1}^{N}|y_i - \hat{y}_i| \quad (13)$$

where $N$ represents the number of data points; $y_i$ is the actual value of electricity price of the training data point indexed by

$i$; and $\hat{y}_i$ is the predicted value of electricity price of the same training data. Note that this metric MAE can also be used on the validation dataset.

The mean squared error (MSE) measures the average squared difference between actual and predicted outputs [36]. The goal of training is to minimize MSE via back propagation which will provides best estimator. The MSE is defined as:

$$MSE = \frac{1}{N}\sum_{i=1}^{N}(y_i - \tilde{y}_i)^2 \qquad (14)$$

## V. SIMULATION RESULTS

The proposed forecast model is tested against the Texas electricity market, ERCOT, to validate its effectiveness and efficiency. This paper employs the hourly settlement point hub price series of the Houston region of ERCOT as a test example of our model [26].

The ERCOT electricity market covers most of the Texas state, with four regional market zones comprising of coast (Houston), west, north, and south of Texas as sub-regions. Bids are submitted by participating companies in interval basis; ERCOT then optimally schedules the generating units based on economics and reliability to meet the system demand while observing resource and transmission constraints and it determines the locational marginal price that is the wholesale electricity price.

The electricity price data originally collected from ERCOT are its real-time market data that are settled down with 15-minute time interval. In this study, hourly resolution is used. The electricity price per hour is calculated by averaging the prices over four consecutive 15-minute dispatch in the same hour. The COAST (Houston) region in ERCOT electricity market is employed in the study. The electric energy price, demand data and temperature data covering the period from January 2015 to December 2018 are used for training process and the data for 2019 are used as the test dataset.

In this study, a total of 34,637 samples were collected, the entire dataset is first divided into two subsets: 31,173 samples (90%) for training and 3,464 samples (10%) for validation. The initial learning rate is set as 0.01, and a learning rate schedule is applied in the training process by reducing the learning rate accordingly, the factor by which the learning rate will be reduced is set as 0.1 and the patience value is set 50 epochs. To leverage the fast-computing abilities of Keras, the machine learning model was trained on NVIDIA RTX 3070 GPUs.

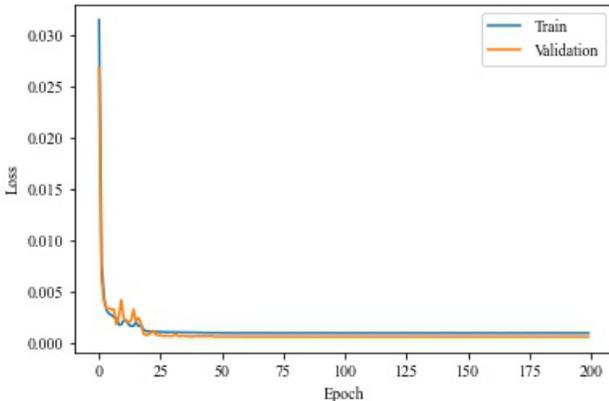

Fig. 8. MSE loss of the proposed ILRCN model with the number of epochs.

Fig. 8 shows the MSE of the training and validation data sets. It can be observed that the MSE is decreasing with the increase in number of epochs. It should be noticed that the MSE stops decreasing at around epoch 200, and hence the training was stopped to obtain the best model.

The prediction accuracies for the proposed ILRCN model and all other models/methods are constructed for different error thresholds. Table I shows the accuracy percentages for hour-ahead and day-ahead forecasting using the naive method explained in Section III for the test dataset under study. With a tolerance of $3, the hour-ahead forecasting has a validation accuracy of 71.55% while the day-ahead forecasting achieves a much lower accuracy of 46.61%, which implies that the hour-ahead forecasting outperforms the day-ahead forecasting by a large margin. This is as expected since the naive method simply uses current price as the future price and works fine only for predictions over very short look-ahead periods.

Table I
Accuracy percentage using naive method for the hour-ahead forecasting and day-ahead forecasting

| Forecasting Error Threshold | Accuracy Percentage for Hour-Ahead Forecasting | Accuracy Percentage for Day-Ahead Forecasting |
|---|---|---|
| $3 | 71.55% | 46.61% |
| $2 | 60.96% | 34.23% |
| $1 | 39.78% | 18.22% |

Table II
Accuracy percentage comparison between FCNN model, LSTM model, LRCN model and the proposed ILRCN model for hour-ahead forecasting

| Forecasting Threshold | Hour-ahead Forecasting Accuracy Percentage | | | | |
|---|---|---|---|---|---|
| | SVR Model | FCNN Model | LSTM Model | LRCN Model | ILRCN Model |
| $3 | 41.82% | 47.50% | 84.32% | 84.40% | 85.55% |
| $2 | 30.66% | 33.77% | 79.77% | 79.67% | 81.26% |
| $1 | 16.31% | 18.13% | 64.98% | 65.92% | 71.70% |

Table III
Accuracy percentage comparison between FCNN model, LSTM model, LRCN model and the proposed ILRCN model for day-ahead forecasting

| Forecasting Threshold | Day-Ahead Forecasting Accuracy Percentage | | | | |
|---|---|---|---|---|---|
| | SVR Model | FCNN Model | LSTM Model | LRCN Model | ILRCN Model |
| $3 | 41.82% | 42.23% | 71.60% | 71.36% | 74.41% |
| $2 | 30.66% | 30.09% | 56.11% | 54.56% | 56.26% |
| $1 | 16.31% | 18.55% | 30.47% | 28.89% | 35.05% |

Table II compares the results of an SVR model, an FCNN model, an LSTM only model, and the proposed LRCN model without the proposed conditional error correction term and the proposed ILRCN model with conditional error correction for the same selected test data for hour-ahead forecasting. It is worth noting that the proposed ILRCN model has the highest validation accuracy of 85.55% with a tolerance of $3, after training of 200 epochs.

Table III compares the results of an SVR model, an FCNN model, an LSTM only model, the proposed LRCN and ILRCN models for the same selected test data for day-ahead forecasting. The LRCN model has very similar validation accuracy with the LSTM model, while the proposed ILRCN model has an accuracy of 74.41% which outperforms all other models.

Comparing Tables II and III, it is observed that the SVR model provides the same results no matter it predicts day-ahead or hour-ahead; this is because SVR model does not capture the time series information and considers all samples independently. This is similar for FCNN while FCNN can improve its prediction accuracy slightly by retraining it with additional information of the operating day.

From Table I and Table III, we can observe that for both hour-ahead and day-ahead forecasting of wholesale electricity price, the proposed ILRCN model outperforms the SVR model, FCNN model, LSTM model and LRCN model by a large margin, as well as the naive method explained in Section III. This demonstrates the efficacy of the proposed ILRCN model for hour-ahead and day-ahead electricity price prediction. It is interesting to observe that the SVR and FCNN models cannot match the naive method while all other neural network based models perform much better than the naive method in both hour-ahead prediction and day-ahead prediction.

The effectiveness of the proposed conditional error correction term can be demonstrated by comparing the performances of the proposed LRCN model and the proposed ILRCN model. When using a threshold of $3, the proposed ILRCN model slightly outperforms the LRCN model by 1.15% for hour-ahead prediction or by 3.05% for day-ahead prediction. However, when applying a threshold of $1, the proposed ILRCN model can achieve much better accuracy by a large margin of 5.78% for day-ahead prediction or 6.16% for hour-ahead prediction.

Table IV
Performance metrics of different models for hour-ahead forecasting

| Performance Metrics | Hour Ahead Forecasting | | | | |
|---|---|---|---|---|---|
| | SVR Model | FCNN Model | LSTM Model | LRCN Model | ILRCN Model |
| MAE | 0.3055 | 0.1362 | 0.0326 | 0.0265 | 0.0254 |
| MSE | 0.3110 | 0.0562 | 0.0020 | 0.0017 | 0.0015 |
| Computation Time (seconds) for Training | 231 | 1,286 | 14,890 | 2,246 | |

Table V
Performance metrics of different models for day-ahead forecasting

| Performance Metrics | Day Ahead Forecasting | | | | |
|---|---|---|---|---|---|
| | SVR Model | FCNN Model | LSTM Model | LRCN Model | ILRCN Model |
| MAE | 0.3055 | 0.3081 | 0.2022 | 0.2059 | 0.1981 |
| MSE | 0.3110 | 0.2933 | 0.1478 | 0.1521 | 0.1355 |
| Computation Time (seconds) for Training | 231 | 1,417 | 15,168 | 2,279 | |

The performance metrics like absolute error of the forecasting results, mean absolute error for the training process and total training time are also computed for various forecasting models and the results are presented in Tables IV and V. It can be observed that the MAE and MSE of the proposed ILRCN model are 0.0254 and 0.0015 respectively for hour-ahead forecasting, which are the lowest among all models. Though it has been demonstrated that the training time for the proposed ILRCN model is comparatively more for the hour-ahead case, the prediction accuracy percentage is higher and absolute error is considerably lower, which indicates the effectiveness of the proposed ILRCN model. It is interesting to note that for day-ahead forecasting LRCN and ILRCN model, with CNN layer included, the computation time for training is reduced considerably while the accuracy increases slightly when compared to the LTSM model, which proves the efficiency of the proposed LRCN architecture. The training time for the LRCN model and ILRCN model is the same, because the conditional error correction term included in the proposed ILRCN model is only used in prediction (not used in training) to mitigate the imperfectness of the trained LRCN model.

Fig. 9 shows the comparison between the predicted settlement point price by LRCN model and the actual price for a subset of the test data under consideration. It is evident from Fig. 9 that the LRCN model does not completely track the variation in settlement point price especially for price spikes. The hour-ahead forecasted settlement point prices by the proposed ILRCN model and the actual prices are shown in Fig. 10 that covers the same time period with Fig. 9; it can be observed that the proposed ILRCN model can very well track the price variation and even the price spikes.

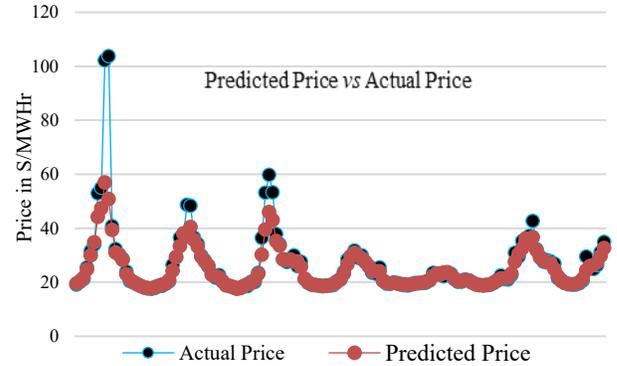

Fig. 9. Hour-ahead forecasted settlement point price by the LRCN model vs. the actual price.

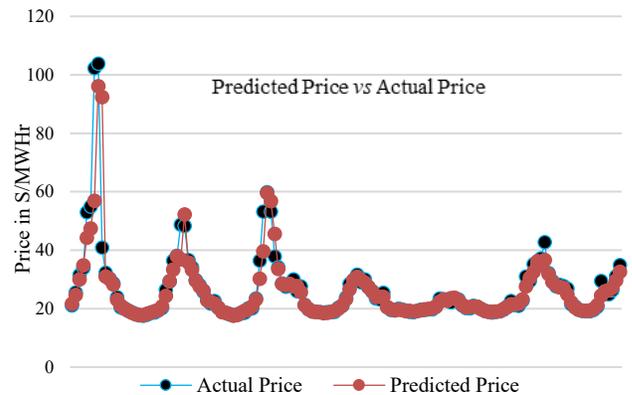

Fig. 10. Hour-ahead forecasted settlement point price by the proposed ILRCN model vs. the actual price.

To summarize, the proposed ILRCN model can achieve high accuracy and low error of real-time wholesale power energy market electricity price prediction than all other models studied in this paper that include a naive method, an SVR model, an FCNN model, an LSTM model and an LRCN model. Although the LSTM model can also obtain a decent accuracy, its training time is about 7 times more than the LRCN and ILRCN models, which demonstrates the efficiency of the architecture of the proposed LRCN model. Note that the above results are achieved only with the use of demand, date, and temperature information as input features. Due to limited data access, the input dataset does not contain any details on generation capacity, electrical network, bidding information, and availability of distributed energy resources in the network that also contribute to settlement point price. Moreover, the electricity price we predict is real-time market price that is much more volatile than day-ahead market price; and persistent volatility consequently results in reduced accuracy for electricity price forecasting. This implies that the proposed ILRCN model may achieve even better results when predicting electricity price of the day-ahead market.

## VI. CONCLUSIONS

This paper proposes an ILRCN model to predict wholesale market electricity price. Various dominant factors like demand profile and temperature contributing to wholesale market price are considered as input features for electricity price forecasting. The proposed ILRCN model for wholesale market electricity price forecasting outperforms the SVR model, FCNN model, LSTM model, and the LRCN model, as well as the naive method in terms of various metrics including training efficiency, accuracy, MSE and MAE, as demonstrated in the case studies section.

The proposed ILRCN model predicts the electricity price with high accuracy and low error both hour-ahead and day-ahead as compared to the naive method, the SVR, FCNN, LSTM and LRCN models. The proposed conditional error correction term can further improve the prediction performance of the proposed LRCN model. In summary, the proposed ILRCN model outperforms both the traditional models and other neural network based models; the proposed ILRCN model is proven to be an accurate and efficient model in settlement point price forecasting. The practicality and feasibility of the proposed ILRCN model are confirmed from the performance metrics. Accurate price prediction also benefits the utility companies to formulate their long-term strategies. With minor adjustment and tune, the ILRCN model proposed in this paper may also be applied in other fields such as load forecasting and variable renewable generation forecasting.